\documentclass[final]{cvpr}

\usepackage{silence}
\makeatletter
\def\input@path{{../}{./others/}}
\makeatother

\usepackage{times}
\usepackage{epsfig}
\usepackage{textcomp}
\usepackage{graphicx}
\usepackage{amsmath}
\usepackage{mathtools}
\usepackage{amssymb}
\usepackage{multirow}
\usepackage{caption}
\usepackage{makecell} 
\usepackage{subcaption}
\usepackage{tabularx}
\usepackage[pass]{geometry}
\usepackage[acronyms]{glossaries}
\usepackage[utf8]{inputenc}
\usepackage{xr-hyper}
\usepackage{layout}
\graphicspath{{figs/},}

\usepackage[pagebackref,breaklinks=true,colorlinks,bookmarks=false]{hyperref}
\usepackage{cleveref}

\Crefname{equation}{Eq.}{Eqs.}
\Crefname{figure}{Fig.}{Figs.}

\newcolumntype{Y}{>{\centering\arraybackslash}X}
\newcolumntype{Z}{>{\raggedleft\arraybackslash}X}

\newcolumntype{L}{>{\raggedright\arraybackslash$}X<$}
\newcolumntype{M}{>{\centering\arraybackslash$}X<$}
\newcolumntype{N}{>{\raggedleft\arraybackslash$}X<$}

\newcolumntype{A}{>{\small}l}
\newcolumntype{B}{>{\small}c}
\newcolumntype{C}{>{\small}r}

\newcommand{\headline}[1]{\par\noindent\emph{\textbf{#1}}~}

\makeatletter
\newcommand*{\addFileDependency}[1]{
  \typeout{(#1)}
  \@addtofilelist{#1}
  \IfFileExists{#1}{}{\typeout{No file #1.}}
}
\makeatother

\usepackage{xspace}
\makeatletter
\providecommand{\onedot}{\futurelet\@let@token\@onedot}
\providecommand{\@onedot}{\ifx\@let@token.\else.\null\fi\xspace}
\makeatother

\providecommand{\etal}{\emph{et al}\onedot}

\newcommand\blfootnote[1]{%
  \begingroup%
  \renewcommand\thefootnote{}\footnote{#1}%
  \addtocounter{footnote}{-1}%
  \endgroup
}

\newacronym{glidenet}{GlideNet}{Global, Local and Intrinsic based Dense Embedding Network}
\newacronym{fe}{FE}{Feature Extractor}
\newacronym{gfe}{GFE}{Global Feature Extractor}
\newacronym{lfe}{LFE}{Local Feature Extractor}
\newacronym{ife}{IFE}{Instance Feature Extractor}

\newacronym{car}{CAR}{Cityscapes Attributes Recognition}
\newacronym{vaw}{VAW}{Visual Attributes in the Wild}

\def\cvprPaperID{11046} 
\def\confYear{CVPR 2022}

\usepackage{background}

\backgroundsetup{
  color=black,
  angle=0,
  opacity=0.7,
  contents={\fontsize{1}{1}\selectfont To appear in the IEEE/CVF CVPR 2022 Proceedings - New Orleans, Louisiana, USA},
  placement=top,
}

\begin{document}

\title{
GlideNet: Global, Local and Intrinsic based Dense Embedding NETwork \\
for Multi-category Attributes Prediction
}

\author{Kareem~Metwaly$^{1}$ \qquad Aerin~Kim$^2$ \qquad Elliot~Branson$^2$ \qquad Vishal~Monga$^1$\\
\hfill$^{1}$The Pennsylvania State University\hfill
$^2$Scale AI\hfill\mbox{}\\
{\tt\small $^1$\{kareem, vum4\}@psu.edu \qquad $^2$\{aerin.kim, elliot.branson\}@scale.com}}

\maketitle

\begin{abstract}

Attaching attributes (such as color, shape, state, action) to object categories is an important computer vision problem. Attribute prediction has seen exciting recent progress and is often formulated as a multi-label classification problem. Yet significant challenges remain in: 
1) predicting a large number of attributes over multiple object categories, 2) modeling category-dependence of attributes, 3) methodically capturing both global and local scene context, and 4) robustly predicting attributes of objects with low pixel-count.  To address these issues, we propose a novel multi-category attribute prediction deep architecture named GlideNet, which contains three distinct feature extractors.
A global feature extractor recognizes what objects are present in a scene, whereas a local one focuses on the area surrounding the object of interest. 
Meanwhile, an intrinsic feature extractor uses an extension of standard convolution dubbed Informed Convolution to retrieve features of objects with low pixel-count utilizing its binary mask. GlideNet then uses gating mechanisms with binary masks and its self-learned category embedding to combine the dense embeddings. Collectively, the Global-Local-Intrinsic blocks comprehend the scene's global context while attending to the characteristics of the local object of interest. 
The architecture adapts the feature composition based on the category via category embedding. Finally, using the combined features, an interpreter predicts the attributes, and the length of the output is determined by the category, thereby removing unnecessary attributes.
GlideNet can achieve compelling results on two recent and challenging datasets -- VAW and CAR -- for large-scale attribute prediction. For instance, it obtains more than $5\%$ gain over state of the art in the mean recall (mR) metric. GlideNet's advantages are especially apparent when predicting attributes of objects with low pixel counts as well as attributes that demand global context understanding. Finally, we show that GlideNet excels in training starved real-world scenarios.
\vspace{-5pt}\protect\blfootnote{\hspace{-10pt}\scriptsize{more info at \url{http://signal.ee.psu.edu/research/glidenet.html}}}
\end{abstract}

\section{Introduction}
To fully comprehend a scene, one should not only be able to detect the objects in the scene but also understand the attributes (properties) of each object detected. Even if two objects belong to the same category, their behavior might vary depending on their attributes. For example, we can't predict the route of a driving vehicle based on a still 2D image alone, unless we know the vehicle's heading/direction and if the vehicle is parked or not.
Accurate classification of objects and their attributes is critical in numerous applications of computer vision and pattern recognition such as autonomous driving where a thorough grasp of the surroundings is essential for safe driving decisions. In order to drive safely, a driver must be able to predict numerous crucial aspects. They include, among other things, the activities of other drivers and pedestrians, the slipperiness of the road surface, the weather, traffic signs and their contents, and pedestrian behavior.

Attributes are often defined as semantic (visual) descriptions of objects in a scene. An object's semantic information includes how it looks (color, size, shape, etc.), interacts with surroundings, and behaviors. The category of an object, in general, determines the set of possible attributes that it can have. For instance, a table might have attributes related to shape, color, and material. However, a human will have a more complicated set of attributes related to age, gender, and activity status (sitting, standing, walking, etc.). Some properties, such as the visible proportion of an object, may exist across multiple categories. Therefore, to accurately predict an object's attributes, we must consider the following:  1) some attributes are unique to certain categories, 2) some categories may share the same attribute, 3) some attributes require a global understanding of the entire scene and 4) some attributes are inherent to the object of interest. In this paper, we present a new algorithm -- \gls{glidenet} -- to tackle the attribute prediction problem. \gls{glidenet}  is capable of addressing the aforementioned listed concerns while also predicting a variety of categories.

Earlier methods for object detection and classification relied heavily on tailored or customized features that are either generated by ORB \cite{rublee2011orb}, SIFT \cite{lowe2004distinctive}, HOG \cite{dalal2005histograms} or other descriptors. Then, the extracted features pass through a statistical or learning module -- such as CRF\cite{lafferty2001conditional} -- to find the relation between the extracted features from the descriptor and the desired output. Recently, Convolutional Neural Networks (CNN) have proven their capability in extracting better features that ease the following step of classification and detection. This has been empirically proven in various fields, such as in object classification \cite{li2020group, huang2019convolutional}, object detection \cite{he2017mask, redmon2016yolo} and inverse image problems such as dehazing \cite{metwaly2020nonlocal, zhang2021learning}, denoising \cite{liu2021invertible,ren2021adaptive}, HDR estimation \cite{liu2020single, metwaly2020attention, chen2021hdrunet}, etc. Deep learning with CNN typically requires a large amount of data for training and regularization \cite{cabon2020vkitti2, yu2020bdd100k, ancuti2020ntire, pougue2021debagreement, caesar2020nuscenes}. Classical methods \cite{antwarg2012attribute, fang2010dangerous} for predicting attributes may require less data, however they perform worse than deep learning based techniques.

In this work, we present a new deep learning approach \gls{glidenet} for attributes prediction that is capable of incorporating problem (dataset) specific characteristics. Our main contributions can be summarized as follows:
\begin{itemize}
    \item We employ three distinct feature-extractors; each has a specific purpose. \gls{gfe} captures global information, which encapsulates information about different objects in the image (their locations and category type). \gls{lfe} captures local information, which encapsulates information related to attributes of the object as well as its category and binary mask. Lastly, \gls{ife} encapsulates information about the intrinsic attributes of objects. It ensures that we estimate characteristics solely from the object's pixels, excluding contributions from other pixels.

    \item We use a novel convolution layer (named Informed Convolution) in the \gls{ife} to focus on intrinsic information of the object related to the attributes prediction.

    \item To learn appropriate weights for each \gls{fe}, we employ a self-attention technique. Utilizing binary mask and a self-learned category embedding, we generate a ``Description'' Then we use a gating mechanism to fine-tune each feature layer's spatial contributions.

    \item We employ a multi-head technique for the final classification stage for two reasons. First, it ensures that the final classification step's weights are determined by the category. Second, the length of the final output can vary depending on the category. This is significant since not every category has the same set of attributes.
\end{itemize}

The term ``class'' can be confusing because it can refer to the object's type (vehicle, pedestrian, etc.) or the value of one of the object's attributes (parked, red, etc). As a result, we avoid using the term ``class'' throughout the work. We use the word ``category'' to refer to the object's type and the word ``attribute'' for one of the semantic descriptions of that object. In addition, we use uppercase letters $X$ to denote images or 2D spatial features, lowercase bold letters $\mathbf{x}$ for 1D features, and lowercase non-bold letters $x$ for scalars, a hat accent over a letter $\hat{x}$ to denote an estimated value and calligraphic letters $\mathcal{X}$ to denote either a mathematical operation or a building block in \gls{glidenet}'s architecture.


\section{Related Work}

Attributes prediction shares common background with other popular topics in research such as object detection \cite{wang2021end, joseph2021towards}, image segmentation \cite{huynh2021progressive, li2021semantic} and classification \cite{liu2021ntie, srinivas2021bottleneck}. 
However, visual attributes recognition has its unique characteristics and challenges that distinguish it from other vision problems such as multi-class classification \cite{reese2020lbcnn} or multi-label classification \cite{durand2019learning, chen2019multi}. 

Examples of these challenges are the possibly large number of attributes to predict, the dependency of attributes on the category type, and the necessity of incorporating both global and local information effectively. This has motivated several past studies to investigate how we could tailor a recognition algorithm that can predict the attributes.

So far, the majority of relevant research has concentrated on a small number of generic attributes. \cite{kalayeh2021symbiosis, wang2017joint, tay2019aanet, rothe2015dex, li2016human, wang2021pedestrian} or a targeted set of categories \cite{he2017adaptively, park2018attribute, yang2020hierarchical, tang2019improving, li2018landmark, abdulnabi2015multi}. For instance, \cite{huo2016vehicle, sun2019vehicle} predict the attributes related to the vehicles. \cite{sun2019vehicle} have proposed a vehicle attributes prediction algorithm. The proposed method uses two branches one to predict the brand of the vehicle and another to predict the color of the vehicle. They use a combined learning schedule to train the model on both types of attributes. Huo \etal \cite{huo2016vehicle} use a convolution stage first to extract important features, then they use a multi-task stage which consists of a fully connected layer per an attribute. The output of each fully connected layer is a value describing that particular attribute. For more details about recent work in vehicle attribute prediction, Ni and Huttunen \cite{ni2021vehicle} have a good survey of recent work, and some existing vehicle datasets for vehicle attributes recognition (e.g. color, type, make, license plate, and model) can be found in \cite{yang2015large, liu2016deep}.

\begin{figure*}[ht!]
    \centering
    \includegraphics[width=\textwidth]{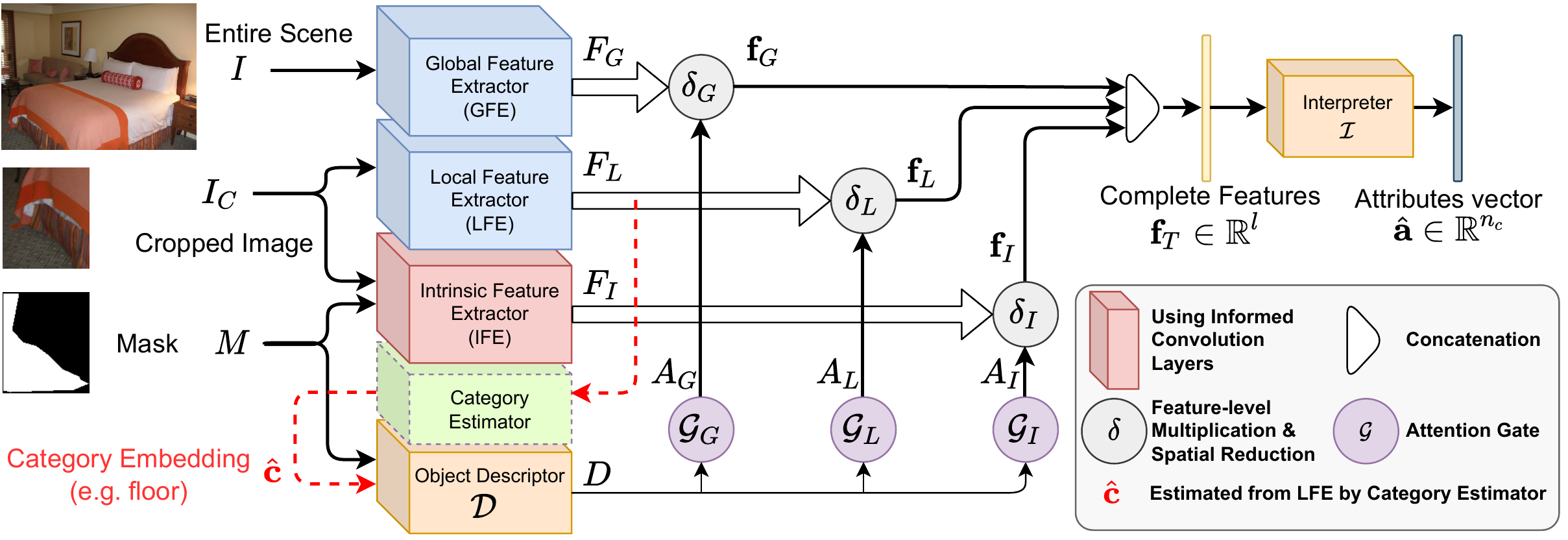}
    \vspace{-20pt}
    \caption{\gls{glidenet} -- the inputs are the image, the binary mask and the category of an object. The output is the attributes of the object. Note that, the category embedding is self-learned from the extracted features of \gls{lfe} using the category estimator. All shown images are taken from \gls{vaw} Dataset.}
    \label{fig:network_architecture}
    \vspace{-13pt}
\end{figure*}

On the other hand, \cite{abdulnabi2015multi, jia2020rethinking, tang2019improving} tackle the prediction of attributes related to pedestrians or humans. Jahandideh \etal \cite{jahandideh2018physical} attempts to predict physical attributes such as age and weight. They use a residual neural network and train it on two datasets; CelebA \cite{liu2015faceattributes} and a self-developed one \cite{liu2015faceattributes}. Abdulnabi \etal \cite{abdulnabi2015multi} learns semantic attributes through a multi-task CNN model, each CNN generates attribute-specific feature representations and shares knowledge through multi-tasking. They use a group of CNN networks that extract features and concatenate them to form a matrix that is later decomposed into a shared features matrix and attribute-specific features matrix. \cite{zhang2020solving} attempt to focus on datasets with missing labels and attempt to solve it with ``background replication loss''. Multiple datasets focus on attributes of humans, but the majority target facial attributes such as eye color, whether the human is wearing glasses or not, $\cdots$, etc. Examples of datasets for humans with attributes are CelebA \cite{liu2015faceattributes} and IMDB-WIKI \cite{rothe2015dex}. Li \etal \cite{li2019visual} propose a framework that contains a spatial graph and a directed semantic graph. By performing reasoning using the Graph Convolutional Network (GCN), one graph captures spatial relations between regions, and the other learns potential semantic relations between attributes.

Only a handful of published work tackled a large set of attributes from a large set of categories \cite{pham2021learning, huang2020image, sarafianos2018deep, yang2020hierarchical}. Sarafianos \etal \cite{sarafianos2018deep} proposed a new method that targeted the issue of class imbalance. Although they focused on human attributes, their method can be extended to other categories as well. Pham \etal \cite{pham2021learning} proposed a new dataset \gls{vaw} that is rich with different categories where each object in an image has three sets of positive, negative, and unlabeled attributes. They use GloVe \cite{pennington2014glove} word embedding to generate a vector representing the object's category.


\vspace{-5pt}
\section{Proposed Model}
\vspace{-5pt}

Universal semantic (visual) attribute prediction is a challenging problem as some attributes may require a global understanding of the whole scene, while other attributes may only need to focus on the close vicinity of the object of interest or even intrinsically in the object regardless of other objects in the scene. We also aspire to estimate the possible attributes of various types of categories. This necessitates a hierarchical structure where the set of predicted attributes depends on the category of the object of interest. In this section, we discuss the details of \gls{glidenet} and the training procedure to guide each \gls{fe} to achieve its purpose.

\subsection{\gls{glidenet}'s Architecture}\label{subsec:net_arch}

\Cref{fig:network_architecture} shows \gls{glidenet}'s network architecture at inference. The input to the model is an image capturing the entire scene ($I$), the category ($C$), and the binary mask ($M$) of the object of interest. The output of the model is a vector ($\textbf{a}$) representing different attributes of that object. \Cref{fig:network_architecture} shows an example where the object of interest is the small portion of the floor below the bed. The output is a vector of the attributes of the floor. We can decompose the information flow in \gls{glidenet} into three consecutive steps; feature extraction, feature composition, and finally interpretation. In the next few subsections, we discuss the details of each step. However, the reader can refer to \Cref{sec:supp_net_arch} for exact numerical values of the parameters of the architecture.

\subsubsection{Feature Extraction}\label{subsubsec:feat_ext}

Feature extraction generates valuable features for the final classification step. It is of utmost importance to extract features that help in predicting attributes accurately. Some of which require an understanding of the whole image while others are intrinsic to the object. In addition, we are interested in the multi-category case. Thus, we need to strengthen the feature extraction process to deal with arbitrary shapes for the object of interest.  For these reasons, we have three \glspl{fe}; namely \acrfull{gfe}, \acrfull{lfe} and \acrfull{ife}. Each \gls{fe} has a specific purpose so that collectively we have a complete understanding of the scene while giving attention to the object of interest.  

\headline{\gls{gfe}} generates features related to the entire image $I$. It produces features that are used for the identification of the most prominent objects in the image. Specifically, the generated features from \gls{gfe} describe objects detected in the image (their center coordinates, their height and width, and their category). We use the backbone of ResNet-50 \cite{he2016deep} network here. We extract features at three different levels of the backbone network to enrich the feature extraction process and for enhanced detection of objects at multiple scales. We denote the extracted features by \gls{gfe} as $F_G^1, F_G^2, F_G^3$ and collectively by $F_G$. Since the extracted features will have different spatial dimensions, we upsample $F_G^2, F_G^3$ to the spatial size of $F_G^1$; which is denoted by $h\times w$ for the height and width, respectively. Let $\mathcal{U}(X, Y)$ represent a function that upsamples $X$ to the spatial size of $Y$ and $\mathcal{S}$ be a concatenation layer, then
\begin{equation}
    F_G = \mathcal{S}\left(F_G^1, \mathcal{U}\left(F_G^2, F_G^1\right), \mathcal{U}\left(F_G^3, F_G^1\right)\right)
    \label{eq:fg}
\end{equation}

\headline{\gls{lfe}} generates features related to the object of interest, but it also considers the object's edges as well as its vicinity. The extracted features from \gls{lfe} are used for the identification of the object's binary mask as well as its category and attributes. \gls{lfe} should be capable of estimating a significant portion of attributes as it focuses on the object of interest in contrast to \gls{gfe}. However, \gls{gfe} is still necessary for some attributes, which require an understanding of other objects in the scene as well. To illustrate, consider a vehicle towing another one. We cannot recognize the attribute ``towing'' without recognizing the existence of another vehicle and their mutual interaction. That is why we employ \gls{gfe} in features extraction. Similar to \gls{gfe}, we use ResNet-50 as the backbone for \gls{lfe}. The extracted features are denoted by $F_L^1, F_L^2, F_L^3$ and collectively by $F_L$. $F_L^2, F_L^3$ are up-sampled to the spatial size of $F_L^1$.
\begin{equation}
    F_L = \mathcal{S}\left(F_L^1, \mathcal{U}\left(F_L^2, F_L^1\right), \mathcal{U}\left(F_L^3, F_L^1\right)\right)
    \label{eq:fl}
\end{equation}

\headline{\gls{ife}} generates intrinsic features of the object of interest, utilizing its binary mask using a novel convolutional layer dubbed as Informed Convolution. It is of great importance to differentiate and distinguish between the objectives of \gls{lfe} and \gls{ife}. Both of them attempt to extract features that predict the object's attributes. However, \gls{ife} generates features related to the intrinsic properties of the object (its texture as an example). On the other hand, \gls{lfe} generates features associated with its neighborhood and the boundaries of the object. To clarify, assume we want to predict the attributes of a pole in an image. \gls{lfe} cannot estimate its color, as typically poles have low aspect ratios; its height is much larger than its width. Thus, the number of pixels contributing to the pole's color is small compared to the total number of pixels in the cropped image $I_C$. Therefore, any typical \gls{fe} will obscure the pole's pixels with other pixels in the cropped image, even if we use an attention scheme to the output features. On the other hand, \gls{ife} cannot understand the interaction of an object with its vicinity, as it only considers the object's pixels while extracting features. As an example, consider an object's exposure to light. \gls{ife} cannot predict the exposure to light accurately; as that requires comparison with other objects in the vicinity of the pole (a dark-red object may be dark due to its low exposure to light or that it intrinsically has that color). Therefore, \gls{lfe} and \gls{ife} supplement each other for a better estimation of attributes. The structure of \gls{ife} resembles the backbone of ResNet-50 where we replace each convolutional layer with an informed-convolutional one (see \Cref{subsec:informed_conv}). The extracted features are denoted by $F_I^1, F_I^2, F_I^3$ and collectively by $F_I$. $F_I^2, F_I^3$ are also up-sampled to the spatial size of $F_I^1$.
\begin{equation}
    F_I = \mathcal{S}\left(F_I^1, \mathcal{U}\left(F_I^2, F_I^1\right), \mathcal{U}\left(F_I^3, F_I^1\right)\right)
    \label{eq:fi}
\end{equation}

Therefore, we have three different sets of features at the end of the feature extraction step; $F_G, F_L, F_I$. Each of them contains features from three levels (dense embeddings) that are all up-sampled to the same spatial size $h\times w$, which we set to $28\times28$ in our implementation.

\begin{figure}
    \centering
    \begin{subfigure}{\linewidth}
    \includegraphics[width=\textwidth]{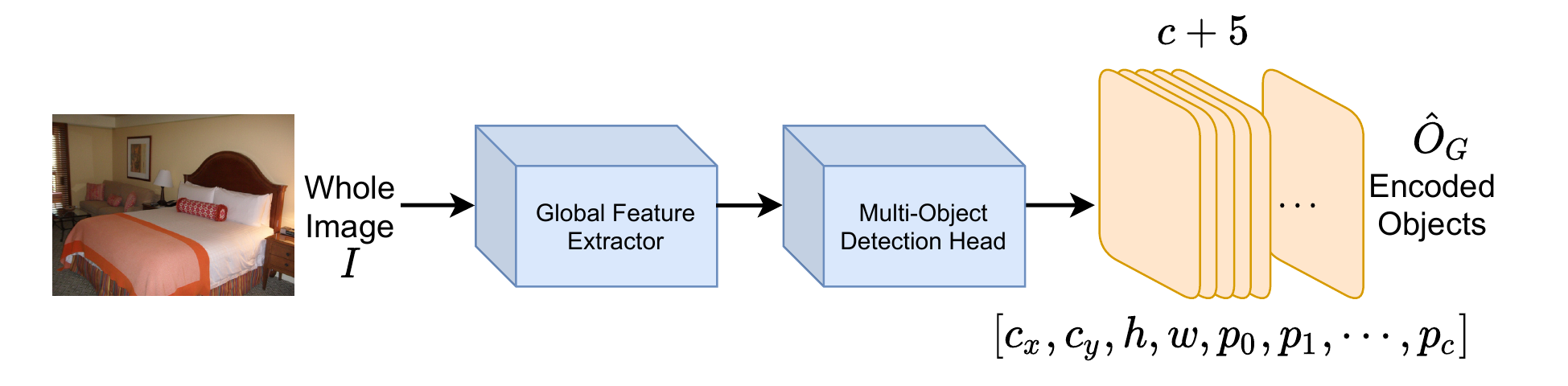}
    \caption{The purpose of \acrshort{gfe} is to understand the scene holistically.}
    \label{fig:gfe_training}
    \end{subfigure}
    \begin{subfigure}{\linewidth}
    \includegraphics[width=\textwidth]{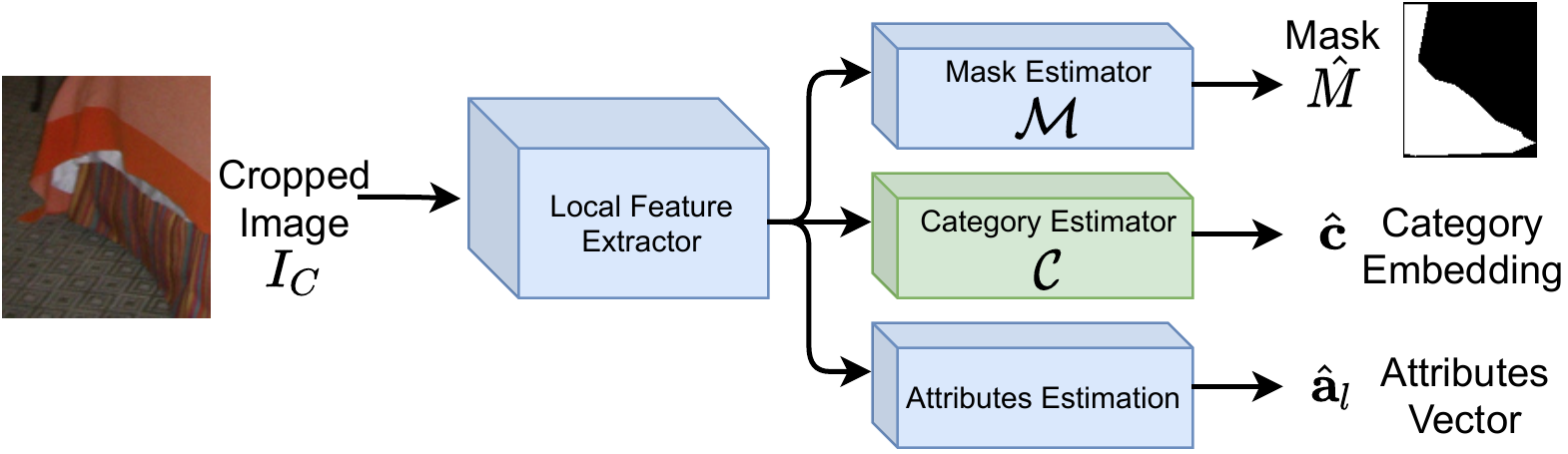}
    \caption{The purpose of \acrshort{lfe} is to extract features related to the object while understanding its vicinity.}
    \label{fig:lfe_training}
    \end{subfigure}
    \begin{subfigure}{\linewidth}
    \includegraphics[width=\textwidth]{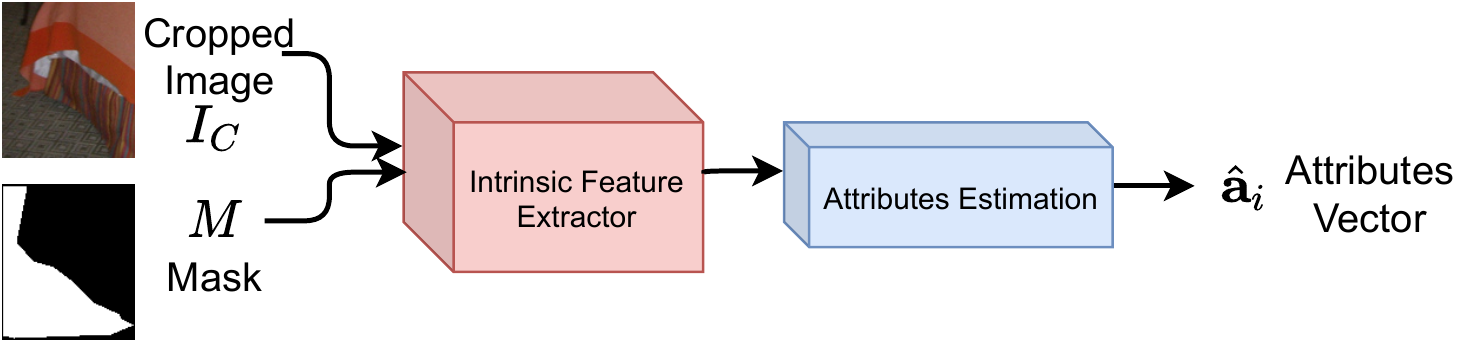}
    \caption{The purpose of \acrshort{ife} is to extract features related to intrinsic properties of the object using Informed Convolution.}
    \label{fig:ife_training}
    \end{subfigure}
    \caption{Training of different feature extractors in Stage I.}
\end{figure}

\begin{figure}
    \centering
    \includegraphics[width=\linewidth]{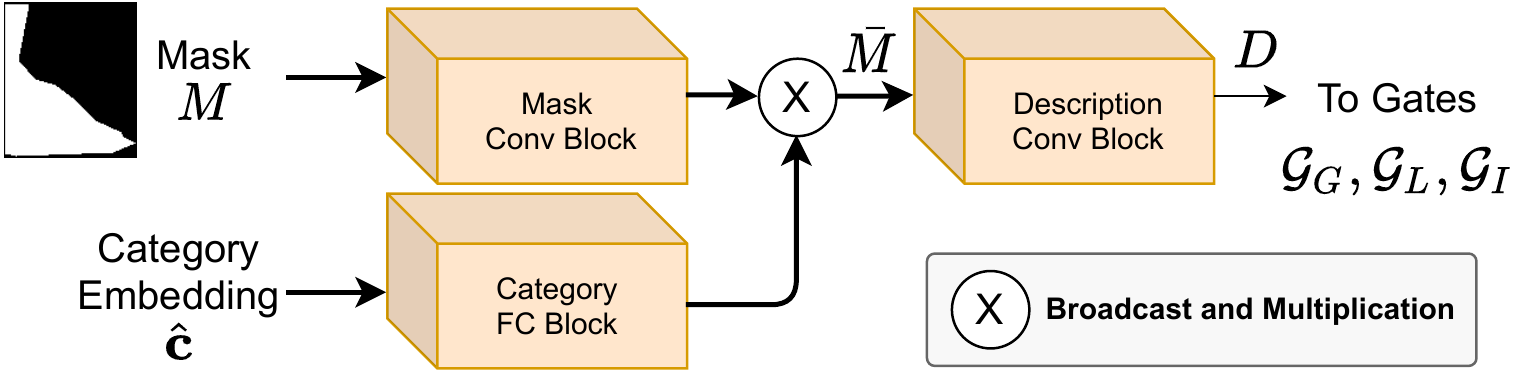}
    \caption{Structure of the Object Descriptor -- the inputs are the binary mask and the self-learned category embedding $\hat{\mathbf{c}}$.}
    \label{fig:object_descriptor}
\end{figure}

\subsubsection{Feature Composition}\label{subsubsec:feat_comp}

Feature composition amalgamates the generated dense embeddings from different feature extractors. A diligent feature composition is indispensable here, as a weak one will impair the extracted features and give all of the attention to only one of the \glspl{fe}. Therefore, we leverage the binary mask of the object of interest besides a self-generated and learnable ``category embedding'' to produce a description $D$ for the composition mechanism. Details about how we generate the ``category embedding'' can be found in \Cref{subsubsec:lfe_training}. After generating the description $D$, it passes by spatial gating mechanisms $\mathcal{G}_G, \mathcal{G}_L, \mathcal{G}_I$ to generate spatial attention weights denoted by $A_G, A_L, A_I$ in \Cref{fig:network_architecture}. Later, we use these weights to reduce the 2D spatial extracted features $F_G, F_L, F_I$ to 1D features $\mathbf{f}_G,  \mathbf{f}_L, \mathbf{f}_I$ through $\delta_G, \delta_L, \delta_I$ , respectively. That effectively generates spatial attention maps to each feature level of each \gls{fe} based on the shape and category of the object. In other words, \gls{glidenet} learns to focus on different spatial locations per each \gls{fe} individually.

The structure of the Object Descriptor ($\mathcal{D}$), \Cref{fig:object_descriptor}, is as follows. First, the binary mask $M$ passes through a convolution block to learn spatial attention based on the object's shape. Meanwhile, the Category Embedding $\hat{\mathbf{c}}$ passes by a fully connected block to learn an attention vector based on the category. Then the category attention vector is broadcasted and multiplied by the mask attention as follows.
\begin{align}
    \Bar{M} &= \hat{\mathbf{c}} \otimes  M\\
    \Bar{M}_i[m,n] &= \hat{c}_i \cdot M_i[m,n]
    \label{eq:obj_desc_intermediate}
\end{align}
where $[m,n]$ represents a spatial location and $i$ represents the channel number. This leads to a composed description for the attention based on the object's shape and category. Finally, a convolution block is used to refine the output description and generates $D$. The exact structure of $\mathcal{D}$ can be found in \Cref{sec:supp_net_arch}.
\begin{equation}
    D = \mathcal{D}\left(M, \hat{\mathbf{c}}\right)
    \label{eq:obj_desc}
\end{equation}

Then, $D$ passes by three different gates $\mathcal{G}_G, \mathcal{G}_L, \mathcal{G}_I$ each has a final Sigmoid activation layer to assert that the output is between $0$ and $1$. Each gate generates a three channels spatial attention map $A$ for each \gls{fe}. Then, $\mathcal{\delta}$ reduces the 2D extracted features from \gls{fe} to 1D features by multiplying each with its corresponding spatial attention map as follows. 
\begin{align}
    A_k &= \mathcal{G}_k\left(D\right), & A_k &\in \mathbb{R}^{3\times h \times w}\\
    f_k &= \delta\left(F_k, A_k\right), & \forall k &\in \{G, L, I\}
\end{align}
\begin{equation}
    \delta(F_k, A_k) \coloneqq  \mathcal{S}_{i=1}^3\left(\sum_{m=1}^{h}\sum_{n=1}^{w} A_k^i[m,n]  F_k^i[m,n]\right)
\end{equation}
where $\mathcal{S}_{i=1}^3(\cdot)$ denotes concatenation for $i\in\{1,2,3\}$ and $F_k^i[m,n]$ represents the generated features of \gls{fe} $k$ at feature level $i$ and spatial location $[m,n]$. Similarly, $A_k^i[m,n]$ is the output attention map from the gate $\mathcal{G}_k$ at feature level $i$ and spatial location $[m,n]$. Finally, the features are combined to get a single 1D feature vector $f_T$ as follows
\begin{equation}
\vspace{-5pt}
    f_T = \mathcal{S}\left(f_G, f_L, f_I\right)
    \label{eq:f_T}
\vspace{-5pt}
\end{equation}

\begin{figure}
    \newcommand{\figwidth}{\linewidth}
    \centering
    \begin{subfigure}[b]{\figwidth}
        \centering
        \includegraphics[width=\textwidth]{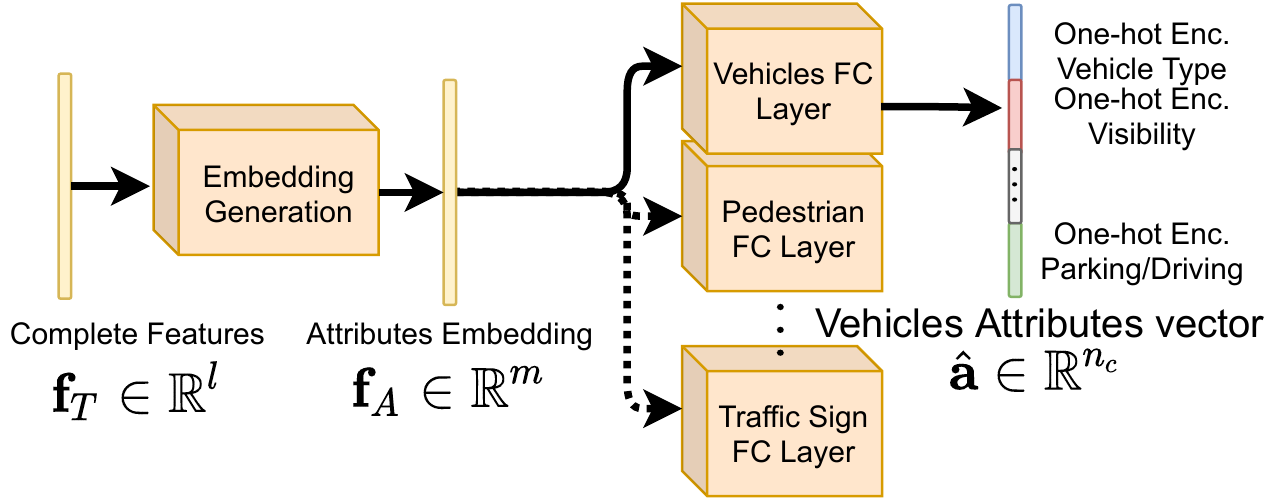}
        \vspace{-1.5em}
        \caption{\acrfull{car} Dataset.}
        \label{fig:car_interpreter}
    \end{subfigure}\hfill
    \begin{subfigure}[b]{\figwidth}
        \centering
        \includegraphics[width=\textwidth]{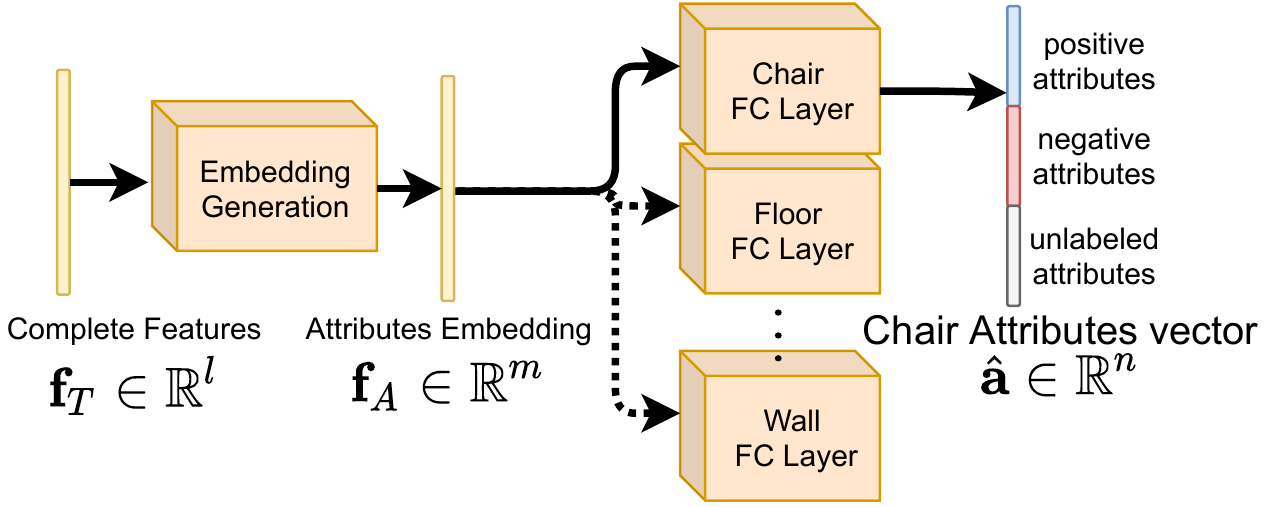}
        \vspace{-1.5em}
        \caption{\acrfull{vaw} Dataset.}
        \label{fig:vaw_interpreter}
    \end{subfigure}
    \vspace{-2em}
    \caption{Structure of the interpreter for different datasets.}
    \label{fig:interpreter}
    \vspace{-10pt}
\end{figure}

\subsubsection{Interpretation}\label{subsubsec:interpretation}

The interpreter translates the final feature vector to meaningful attributes. Its design depends on the final desired attributes outputs. In \Cref{sec:exp_res}, we experiment with two datasets \gls{vaw} and \gls{car}. Both datasets are very recent and focus on a large set of categories with various possible attributes. However, there are some differences between them. Specifically, \gls{vaw} has three different labels (positive, negative, and unlabeled). On the other hand, \gls{car} doesn't have unlabeled attributes; it has a complex taxonomy where each category has its own set of attributes, and each attribute has a set of possible values it may take. This obligates the interpreter to depend on the training dataset and the final desired output. 

Therefore, two models are provided in \Cref{fig:interpreter}. In both cases, we first start with a dimensional reduction fully connected layer from $\mathbb{R}^l$ to $\mathbb{R}^m$; $m < l$. That enables us to create multiple heads for each category without increasing the memory size drastically. Then, the reduced features $\mathbf{f}_A$ passes by a single head corresponding to the category of the object of interest. For \gls{car} in \Cref{fig:car_interpreter}, the output size $n_c$ varies from one head to another depending on the taxonomy of category $c$. While for \gls{vaw} in \Cref{fig:vaw_interpreter}, the output size is the same $n = 620$. The other difference between the two interpreters is in the possible values the output can take. In \gls{vaw}, the output ranges from $0$ to $1$, where $0$ represents negative attributes and $1$ represents positive ones (unlabeled attributes are disregarded in training). In \gls{car}, the output is not binary as some attributes have more than two possible values. Therefore, we encode each attribute as one hot encoder. For example, the ``Vehicle Form'' attribute can take one of 11 values such as ``sedan'', ``Van'', etc. Thus, we have a vector of 11 values where ideally we want the value $1$ at the correct form type and $0$ elsewhere. It's noteworthy to mention that \gls{car} has an ``unclear'' value for all attributes. We skip attributes with unclear values during training.

\begin{figure}
    \newcommand{\figwidth}{0.32\linewidth}
    \centering
    \begin{subfigure}[b]{\figwidth}
        \centering
        \includegraphics[width=\textwidth]{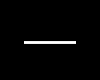}
        \vspace{-1.8em}
        \caption{\scriptsize{Input Mask}}
        \label{fig:input_original_mask}
    \end{subfigure}\hfill
    \begin{subfigure}[b]{\figwidth}
        \centering
        \includegraphics[width=\textwidth]{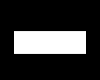}
        \vspace{-1.8em}
        \caption{\scriptsize{Partial Convolution}}
        \label{fig:partial_conv_output_mask}
    \end{subfigure}\hfill
    \begin{subfigure}[b]{\figwidth}
        \centering
        \includegraphics[width=\textwidth]{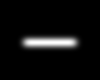}
        \vspace{-1.8em}
        \caption{\scriptsize{Informed Convolution}}
        \label{fig:informed_conv_output_mask}
    \end{subfigure}
    \vspace{-1em}
    \caption{An input mask and its propagated mask after $5$ layers for different update rules - kernel size = $5$.}
    \label{fig:mask_update}
    \vspace{-15pt}
\end{figure}
\subsection{Training}\label{subsec:training}
\vspace{-5pt}

Since \gls{glidenet} has a complex architecture, tailored training of the model is necessary to lead each \gls{fe} to its objective. Therefore, we have developed a customized training scenario for \gls{glidenet} which divides the training into two stages. In \textbf{Stage I}, we focus on guiding \glspl{fe} to a reasonable good status of their objective by adding some temporary decoders to guide the feature extraction process. The objective in Stage I is to have powerful and representative \glspl{fe}. Therefore, we do not train $\mathcal{D}$ nor $\mathcal{I}$ in Stage I. In \textbf{Stage II}, we focus on the actual objective of \gls{glidenet}, which is predicting the attributes accurately. Therefore, we remove the temporary decoders, and we train the whole network structure as in \Cref{fig:network_architecture}. The details of training \glspl{fe} in Stage I is detailed in \Cref{subsubsec:gfe_training,subsubsec:lfe_training,subsubsec:ife_training} while \Cref{subsubsec:stage_2} discusses the training in Stage II.

\vspace{-10pt}
\subsubsection{\acrfull{gfe}}\label{subsubsec:gfe_training}
\vspace{-5pt}

\gls{gfe} is trained as in \Cref{fig:gfe_training} by having a temporary objects decoder that attempts to detect the objects in the input image $I$, their categories and their bounding boxes center locations $(c_x, c_y)$, widths $w$ and heights $h$. $\hat{O}_G$ has $c+5$ channels; $c$ of which are a one-hot representation of the category ($\hat{\mathbf{P}}$), $4$ values for the bounding box, and the remaining value is the probability of having the center of an object in that pixel ($\hat{P}_0$). The training loss term for \gls{gfe} is as follows.
\vspace{-5pt}
\begin{equation}
\vspace{-5pt}
\begin{split}
    \mathcal{L}_\text{g} &= \lambda_{gp0}\mathcal{L}_\text{BCE}\left(P_0, \hat{P}_0\right) + \lambda_{gp}\mathcal{L}_\text{CE}\left(\mathbf{P}, \hat{\mathbf{P}}\right) \\
    &+ \lambda_{gd}\left[\mathcal{L}_\text{MSE}\left(H,\hat{H}\right) + \mathcal{L}_\text{MSE}\left(W,\hat{W}\right)\right] \\ 
    &+ \lambda_{gc}\left[\mathcal{L}_\text{MSE}\left(C_x, \hat{C}_x\right) + \mathcal{L}_\text{MSE}\left(C_y, \hat{C}_y\right)\right]
\end{split}
\label{eq:global_loss_1}
\end{equation}
where $\mathcal{L}_\text{BCE}$ is the Binary Cross Entropy loss, $\mathcal{L}_\text{CE}$ is the multi-class Cross Entropy and $\mathcal{L}_\text{MSE}$ is the Mean-Square-Error loss. $\lambda_{gp0}, \lambda_{gp}, \lambda_{gd}, \lambda_{gc}$ are hyperparameters used to tune the importance of each term.

\begin{table*}
    \centering
    \caption{Comparison Between \gls{glidenet} and other state-of-the-art methods on two challenging datasets \gls{car} and \gls{vaw}}
    \vspace{-1.1em}
    \label{tab:results}
    \begin{tabularx}{\linewidth}{r||Y|Y|Y|Y||Y|Y|Y|Y}
    \Xhline{4\arrayrulewidth}
         \multicolumn{1}{c||}{\multirow{2}{*}{Method}}& \multicolumn{4}{c||}{\acrfull{vaw}\cite{pham2021learning}} & \multicolumn{4}{c}{\acrfull{car}\cite{metwaly2022car}}  \\\cline{2-9}
         & mA & mR & mAP & F1 & mA & mR & mAP & F1 \\\Xhline{3\arrayrulewidth}
        Durand \etal \cite{durand2019learning} & $0.689$ & $0.643$ & $ 0.623$ & $0.632$ & $0.641$ & $0.629$ & $0.637$ & $0.635$ \\\hline
        Jiang \etal \cite{jiang2020defense} & $0.503$ & $0.631$ & $ 0.564$ & $0.597$ & $0.668$ & $0.659$ & $0.671$ & $0.654$ \\\hline
        Sarafianos \etal \cite{sarafianos2018deep} & $0.683$ & $0.647$ & $ 0.651$ & $0.646$ & $0.701$ & $0.699$ & $0.705$ & $0.703$ \\\hline
        Pham \etal \cite{pham2021learning} & $0.715$ & $0.717$ & $ 0.683$ & $0.694$ & $0.731$ & $0.727$ & $0.739$ & $0.720$ \\\hline\hline
        \textbf{\gls{glidenet}} & $\mathbf{0.737}$ & $\mathbf{0.768}$ & $\mathbf{0.712}$ & $\mathbf{0.725}$ & $\mathbf{0.781}$ & $\mathbf{0.802}$ & $\mathbf{0.788}$ & $\mathbf{0.796}$ \\\Xhline{4\arrayrulewidth}
    \end{tabularx}
    \vspace{-1.5em}
\end{table*}

\vspace{-10pt}
\subsubsection{\acrfull{lfe}}\label{subsubsec:lfe_training}
\vspace{-5pt}

\Cref{fig:lfe_training} shows the training of \gls{lfe}. Here, we use three decoders; two temporary decoders for the binary mask $\mathcal{M}$ and attributes, and one decoder for the category embedding $\mathcal{C}$. The training loss term for \gls{lfe} is as follows.
\vspace{-3pt}
\begin{equation}
\footnotesize
\mathcal{L}_\text{l} = \lambda_{lm}\mathcal{L}_\text{BCE}\left(M, \hat{M}\right) + \lambda_{lc}\mathcal{L}_\text{CE}\left(C, \hat{C}\right) + \lambda_{la}\mathcal{L}_\text{BCE}\left(a, \hat{a}\right)
\end{equation}
where $\lambda_{lm}, \lambda_{lc}, \lambda_{la}$ are hyperparameters to tune the importance of each term.

The category embedding encapsulates visual similarities between different categories unlike a word embedding  \cite{pennington2014glove}, which was previously used in \cite{pham2021learning}. We reason that learnable vectors, rather than static pre-trained word embedding, capture greater visual similarities between objects depending on their attributes; a teddy-bear is visually similar (attribute-wise) to a toy more than to an actual real bear.


\vspace{-5pt}
\subsubsection{\acrfull{ife}}\label{subsubsec:ife_training}
\Cref{fig:ife_training} depicts the training of \gls{ife}. It uses Informed Convolution layers detailed in \Cref{subsec:informed_conv} to focus on the intrinsic attributes. Its training loss term is as follows.
\vspace{-5pt}
\begin{equation}
\vspace{-5pt}
    \mathcal{L}_\text{i} = \lambda_{ia}\mathcal{L}_\text{BCE}\left(a, \hat{a}\right)
    \label{eq:ife_loss_1}
\end{equation}
where $\lambda_{ia}$ is a hyperparameter. Therefore, the complete training loss function in Stage I is as follows.
\vspace{-5pt}
\begin{equation}
\vspace{-5pt}
    \mathcal{L}_\text{I} = \mathcal{L}_\text{g} + \mathcal{L}_\text{l} + \mathcal{L}_\text{i}
    \label{eq:total_loss_1}
\end{equation}

\subsubsection{Stage II}\label{subsubsec:stage_2}
In Stage II, the following loss function focuses on generating the final attributes vector correctly from the interpreter while maintaining accurate category embedding $\hat{c}$.
\vspace{-5pt}
\begin{equation}
\vspace{-5pt}
    \mathcal{L}_\text{II} = \mathcal{L}_\text{BCE}\left(a, \hat{a}\right) + \lambda_{lc2}\mathcal{L}_\text{CE}\left(C, \hat{C}\right)
    \label{eq:loss_2}
\end{equation}
Therefore, the main goal is to predict the desired attributes. However, we keep the term for the category embedding to ensure the convergence of the category embedding during training in Stage II.

\subsection{Informed Convolution}\label{subsec:informed_conv}
The utilization of the binary mask in the feature extraction process has been previously applied in image inpainting problems in \cite{liu2018image, yu2019free, chang2019free}. \cite{yu2019free, chang2019free} used learnable gates to find the best mask-update rule, which is not suitable here as we want \gls{ife} to only focus on intrinsic attributes of the object. Therefore, a learnable update rule does not guarantee the convergence to a physically meaningful updated mask. Inspired by \cite{liu2018image} we perform a mask-update rule as follows.
\begin{equation}
    X^{(i+1)} = \left\{
    \begin{array}{ll}
         \frac{k^2\cdot W^T}{\sum M^{(i)}}\left(X^{(i)} \odot M^{(i)}\right)  & \text{if}\max M^{(i)} > 0, \\
        0 & \text{otherwise}
    \end{array}
    \right.
\end{equation}
\begin{equation}
    M^{(i+1)} = \left\{
    \begin{array}{ll}
        \frac{1}{k^2}\sum M^{(i)} & \text{if} \max M^{(i)} > 0, \\
        0 & \text{otherwise}
    \end{array}
    \right.\\
\end{equation}
where $k$ is the kernel size of the convolution layer, $X^{(i)}, M^{(i)}$ are the input features and input binary mask for convolution layer $i$ that is only visible for the kernel and $\odot$ represents element-wise multiplication. It is important to notice the difference between our mask-update rule and the one provided in \cite{liu2018image}. \Cref{fig:informed_conv_output_mask} shows an output example based on the update rule. In our case, each pixel contributes to the new mask by a soft value that depends on the contribution of the object of interest at that spatial location. Furthermore, Informed Convolution can be reduced to a regular convolution if the binary mask was all ones. In this case, the object of interest completely fills the image and the intrinsic features would be any feature that we can extract from the image. It is also noteworthy to recognize the difference between Informed Convolution and Masked Convolution presented in \cite{van2016pixel}, where the authors are interested in generating an image from a caption by using a mask to ensure the generation of a pixel depends only on the already generated pixels. Their purpose and approach is entirely different.


\section{Experiments and Results}\label{sec:exp_res}

In this section, we validate the effectiveness of \gls{glidenet} and provide results of extensive experiments to compare it with existing state-of-the-art methods. Specifically, we provide results on two challenging datasets for attributes prediction -- \gls{vaw} \cite{pham2021learning} and \gls{car} \cite{metwaly2022car}. In addition, we perform several ablation studies to show the importance of various components of \gls{glidenet}. While we can consider other datasets such as \cite{patterson2016coco, krishna2017visual}, they lack diversity in either categories or attributes. However, \gls{vaw} has $260,895$ instances; each with $620$ positive, negative and unlabeled attributes. On the other hand, \gls{car} \cite{car_api} has $32,729$ instances focusing on self-driving. Unlike \gls{vaw}, \gls{car} has a complex hierarchical structure for attributes, where each category has its own set of possible attributes. Some attributes may exist over several categories (such as visibility) and some other are specific to the category (such as walking for pedestrian).

\headline{Experiment setup:} the model is implemented using PyTorch framework \cite{paszke2019pytorch}. We choose the values of $\lambda_{gp0}$, $\lambda_{gp}$, $\lambda_{gd}$, $\lambda_{gc}$, $\lambda_{lm}$, $\lambda_{lc}$, $\lambda_{la}$, $\lambda_{ia}$ and $\lambda_{lc2}$ to be $1, 0.01, 0.5, 0.5, 0.1, 0.01, 1, 1$ and $0.01$, respectively by cross validation \cite{monga2018handbook}. We trained the model for 15 epochs at Stage I and then 10 epochs for Stage II. More details can be found in \Cref{sec:supp_exp_setup}. 

\headline{Evaluation Metrics:} mean balanced Accuracy (mA), mean Recall (mR), F$_1$-score and mean Average Precision (mAP) are used for evaluation. They are unanimously used for classification and detection problems. Specifically, they have been used in existing work for attributes prediction such as \cite{pham2021learning, durand2019learning, sarafianos2018deep, li2017improving, jiang2020defense, anderson2018bottom}. Excluding mAP, we calculate these metrics over each category then compute the mean over all categories. Therefore, the metrics are balanced; a frequent category contributes as much as a less-frequent one (no category dominates any metric). However for mAP, the mean is computed over the attributes similar to \cite{pham2021learning, gupta2019lvis}. We compute the mean over attributes in case of mAP to ensure diversity in metrics used in evaluation. As in this case, we ensure having balance between different attributes. All metrics are defined as follows.
\vspace{-3pt}
$$
\vspace{-3pt}
\text{mA} = \frac{1}{2c}\sum_{i=1}^{c}\frac{\text{TP}_i}{\text{P}_i} + \frac{\text{TN}_i}{\text{N}_i},\quad\quad
\text{F}_1 = \frac{2 \text{mP} * \text{mR}}{\text{mP} + \text{mR}},
$$
$$
\text{mP} =  \frac{1}{c}\sum_{i=1}^{c}\frac{\text{TP}_i}{\text{PP}_i},\quad
\text{mR} =  \frac{1}{c}\sum_{i=1}^{c}\frac{\text{TP}_i}{\text{P}_i},\quad
\text{mAP} = \frac{1}{n}\sum_{j=1}^{n} \text{AP}_j
\vspace{-3pt}
$$
where $c$ and $n$ are the numbers of categories and attributes respectively. TP$_i$, TN$_i$, P$_i$, N$_i$ and PP$_i$ are the number of true-positive, true-negative, positive samples, negative samples and predicted-positive samples for category $i$. AP$_{j}$ is the average of the precision-recall curve of attribute $j$ \cite{lin2014microsoft}. 

Since some attributes are unlabeled in \gls{vaw}, we disregard them in the evaluation as \cite{pham2021learning} did. Conversely, \gls{car} does not contain unlabeled attributes. It has, however, a complex hierarchical taxonomy of attributes that requires modification in the metrics used. For instance, most attributes are not binary. They can take more than two values; a ``visibility'' attribute may take one of five values. Therefore, we define TP and TN per attribute per category. Then we compute the mean over all attributes of all categories. For example, mA would be as follows.
\vspace{-5pt}
\begin{equation*}
\vspace{-5pt}
        \text{mA} = \frac{1}{2c}\sum_{i=1}^{c}\left( \frac{1}{n_i}\sum_{j=1}^{n_i}\frac{\text{TP}_{i,j}}{\text{P}_{i,j}} + \frac{\text{TN}_{i,j}}{\text{N}_{i,j}}\right)
\end{equation*}
where $n_i$ is the number of attributes of category $i$. TP$_{i,j}$ is the positive samples of attribute $j$ of category $i$. Similarly, we can extend the definition of other metrics to suit the taxonomy of \gls{car}. For further details, the reader is encouraged to check \Cref{sec:supp_exp_setup}.

\headline{Results on \gls{vaw} and \gls{car}:} \Cref{tab:results} shows the results of \gls{glidenet} in comparison with four state-of-the-art method over \gls{vaw} and \gls{car}. In \gls{vaw}, \gls{glidenet} obtained better values in all metrics. More prominently, it was able to gain $5\%$ in mR metric than the closest method \cite{pham2021learning}. This is mainly due to \gls{glidenet}'s usage of \gls{ife} and \gls{gfe} to detect attributes requiring global and intrinsic understanding.
In \gls{car}, \gls{glidenet} was capable of achieving even a higher gain ($\sim 8\%$ mR). \gls{glidenet} can be trained directly with \gls{car} dataset due to its varying output length. However, we had to slightly modify the architecture of other method to work with \gls{car}.

\begin{table}
    \caption{Ablation study over dense embeddings}
    \vspace{-1em}
    \label{tab:dense_embeddings_ablation}
    \centering
    \begin{tabularx}{\linewidth}{r||Y|Y|Y|Y}\Xhline{4\arrayrulewidth}
    \multicolumn{1}{c||}{Method} & mA & mR & mAP & F1 \\\Xhline{2\arrayrulewidth}
    \gls{lfe} only & $0.612$ & $0.639$ & $0.620$ & $0.613$ \\\hline
    \gls{lfe}$+$\gls{gfe} & $0.661$ & $0.644$ & $0.671$ & $0.668$ \\\hline
    \gls{lfe}$+$\gls{ife} & $0.719$ & $0.724$ & $0.699$ & $0.705$ \\\hline
    \textbf{\gls{glidenet}} & $\mathbf{0.737}$ & $\mathbf{0.768}$ & $\mathbf{0.712}$ & $\mathbf{0.725}$ \\\Xhline{4\arrayrulewidth}
    \end{tabularx}
    \vspace{-1.2em}
\end{table}
\begin{table}
    \caption{Ablation study over Objects with low pixel count}
    \vspace{-1em}
    \label{tab:informed_conv_ablation}
    \centering
    \begin{tabularx}{\linewidth}{r||Y|Y|Y|Y}\Xhline{4\arrayrulewidth}
    \multicolumn{1}{c||}{Method} & mA & mR & mAP & F1 \\\Xhline{2\arrayrulewidth}
    Pham \etal\cite{pham2021learning} & $0.619$ & $0.655$ & $0.603$ & $0.626$ \\\hline
    \gls{glidenet} w/o \gls{ife} & $0.658$ & $0.691$ & $0.643$ & $0.647$ \\\hline
    \textbf{\gls{glidenet}} & $\mathbf{0.704}$ & $\mathbf{0.721}$ & $\mathbf{0.680}$ & $\mathbf{0.698}$ \\\Xhline{4\arrayrulewidth}
    \end{tabularx}
    \vspace{-1.2em}
\end{table}
\begin{table}
    \caption{Comparison between \gls{glidenet} with and without $\mathcal{D}$}
    \vspace{-1em}
    \label{tab:descriptor_ablation}
    \centering
    \begin{tabularx}{\linewidth}{r||Y|Y|Y|Y}\Xhline{4\arrayrulewidth}
    \multicolumn{1}{c||}{Method} & mA & mR & mAP & F1 \\\Xhline{2\arrayrulewidth}
    \gls{glidenet} w/o $\mathcal{D}$ & $0.720$ & $0.725$ & $0.696$ & $0.708$ \\\hline
    \textbf{\gls{glidenet}} & $\mathbf{0.737}$ & $\mathbf{0.768}$ & $\mathbf{0.712}$ & $\mathbf{0.725}$ \\\Xhline{4\arrayrulewidth}
    \end{tabularx}
    \vspace{-1.2em}
\end{table}
\begin{table}
    \caption{Ablation study over category embedding}
    \vspace{-1em}
    \label{tab:category_ablation}
    \centering
    \begin{tabularx}{\linewidth}{r||Y|Y|Y|Y}\Xhline{4\arrayrulewidth}
    \multicolumn{1}{c||}{Method} & mA & mR & mAP & F1 \\\Xhline{2\arrayrulewidth}
    \gls{glidenet} w/o CE & $0.725$ & $0.731$ & $0.701$ & $0.712$ \\\hline
    \textbf{\gls{glidenet}} & $\mathbf{0.737}$ & $\mathbf{0.768}$ & $\mathbf{0.712}$ & $\mathbf{0.725}$ \\\Xhline{4\arrayrulewidth}
    \end{tabularx}
    \vspace{-1.2em}
\end{table}
\begin{figure}
    \centering
    \includegraphics[width=0.7\linewidth]{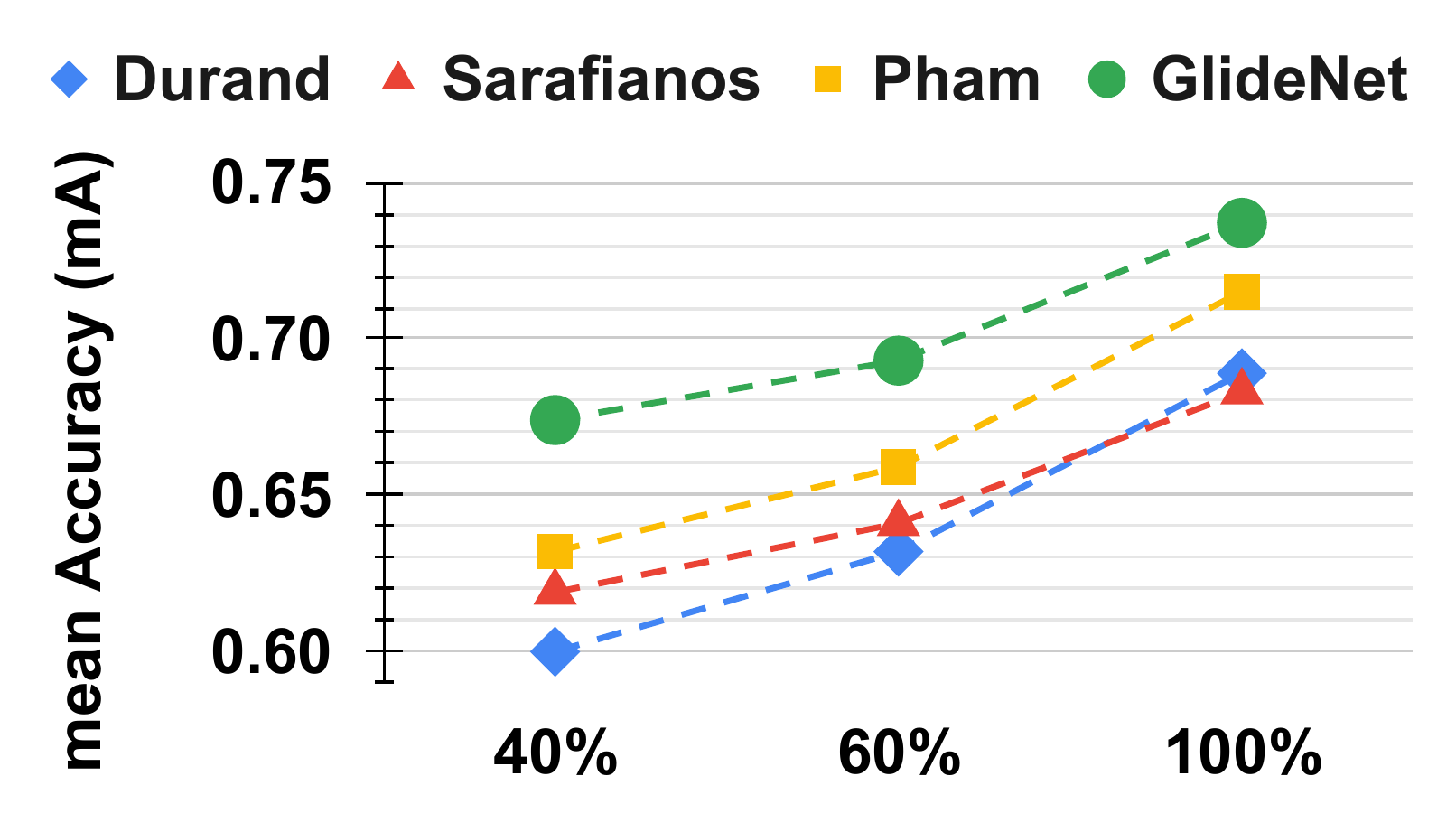}
    \vspace{-10pt}
    \caption{Comparison against training size (VAW Dataset).}
    \label{fig:training_size_ablation}
    \vspace{-1.7em}
\end{figure}

\subsection{Ablation Study}

Several ablation studies are presented here to demonstrate the importance of the unique components in \gls{glidenet}. Only ablations from the VAW dataset are shown here, however, similar behavior was noticed in CAR as well.

\headline{Dense Embedding:} \Cref{tab:dense_embeddings_ablation} shows the results of \gls{glidenet} with different combinations of \glspl{fe}. We achieve best results by using all \glspl{fe}. Notice that the gain from using \gls{ife} is higher than \gls{gfe}. This is expected given most attributes in \gls{vaw} focus on the object of interest itself and do not require a lot of global context. However, \gls{gfe} is still valuable when global understanding of the scene is necessary, as in \gls{car}.

\headline{Informed Convolution:} We retrained the model with a restricted dataset comprising objects with low pixel counts to demonstrate the usefulness of Informed Convolution layers. We specifically identified examples with a lower than $0.35$ ratio between their binary mask and their corresponding bounding boxes. This reflects the goal of Informed Convolution layers, which is to give low-pixel-count objects special attention.

Because the only architectural difference between \gls{ife} and \gls{lfe} is in the usage of Informed Convolution layers, we test two scenarios: one with and one without \gls{ife}. In all measures, \gls{glidenet} obtains the best performance, as seen in \Cref{tab:informed_conv_ablation} by meaningful margin.

\headline{Object Descriptor:} \Cref{tab:descriptor_ablation} shows a comparison between \gls{glidenet} with and without the Object Descriptor $\mathcal{D}$. Despite the fact that the results without $\mathcal{D}$ are less than ideal, they are still meaningfully higher than \cite{pham2021learning}. This suggests that the generated dense embeddings are helping in better attributes recognition. The feature composition of  $\mathcal{D}$, on the other hand, is superior.

\headline{Semantic Embedding:} \gls{glidenet} uses a self-learned category embedding that encapsulate semantic similarities between objects. If the category embedding confuses two categories, it is most likely owing to their visual similarities. In prior studies \cite{pham2021learning}, word embeddings \cite{pennington2014glove} were used to capture the semantic but a word embedding alone would not be sufficient to capture the visual similarities. \Cref{tab:category_ablation} shows a comparison of \gls{glidenet} by swapping the Category Embedding (CE) with GloVe \cite{pennington2014glove} -- a word embedding.

\headline{Limited Training Scenario:} We also perform a limited training data size comparison between \gls{glidenet} and other methods in \Cref{fig:training_size_ablation}. The training data size is limited to $60\%$ and $40\%$ of the original training data size of \gls{vaw} while keeping the validation set as it is. Although all methods suffer in the limited data size scenario, \gls{glidenet} shows a much more graceful decay in comparison to other methods.


\vspace{-5pt}
\section{Conclusion}
\vspace{-5pt}
Global, Local, and Intrinsic based Dense Embedding Network (GlideNet) is a novel attributes prediction model that can work with a variety of datasets and taxonomies of categories and attributes. It surpasses existing state-of-the-art approaches, and we believe this is due to the use of a variety of Feature Extractors (FEs), each with its distinct goal. A two-stage training program establishes their objectives. 

Furthermore, the self-attention method, which combines a binary mask and a self-learned category embedding, fuses dense embeddings based on the object's category and shape and achieves richer composed features. The suggested Informed Convolution-based module estimates attributes for objects in the cropped image that have a very low pixel contribution.
A rigorous ablation study and comparisons with other SOTA methods demonstrated the advantages of GlideNet's unique blocks empirically.


\newpage

\appendix

\section{Introduction of the Supplementary}\label{sec:supp_intro}

This is a supplementary that contains some useful information for accurately reproducing findings, as well as the reasoning for using \gls{vaw} and \gls{car} datasets for experiments in the GlideNet paper.

The complete specifications and setup of the proposed network  -- \gls{glidenet} -- architecture are stored in \Cref{sec:supp_net_arch}.

\Cref{sec:supp_disc_data} discusses the two datasets, \gls{vaw} and \gls{car} datasets, used in the evaluation and why we chose those two datasets in particular. 

All of the configurations for training \gls{glidenet} are contained in \Cref{sec:supp_exp_setup}.

\section{Network Architecture Details}\label{sec:supp_net_arch}
In this section, we discuss the exact details of each building block of \gls{glidenet}. The reader is encouraged to read \Cref{subsec:net_arch} first to understand the purpose of each building block. Here, we only show the configurations without a detailed description of the purpose. 
In our training algorithms (\Cref{subsec:training}), we have two stages. The first stage uses `temporary' decoders that are removed later in both Stage II of training and the inference stage. The details of these temporary decoders are found in \Cref{subsec:supp_mask,subsec:supp_modh,subsec:supp_attributes_1}

\subsection{\acrfullpl{fe}}\label{subsec:supp_fe}
For all \glspl{fe}, we use the backbone of ResNet-50 \cite{he2016deep}. Specifically, we take the output after layers $2, 3, \& 4$ as our features. The inputs to the \glspl{fe} are always resized to $224\times224$. Since the output features don't match in spatial dimensions, we upsample them to the size of the largest, which is $28\times28$. The total number of the output channels for each \gls{fe} in this case is $128+128+512 = 768$. In the case of \gls{ife}, we replace each convolution layer with the novel layer proposed in \Cref{subsec:informed_conv}. However, other than the usage of the (Mask-)Informed Convolution concept, it has the same exact structure as other \glspl{fe}. The output of \gls{gfe}, \gls{lfe} and \gls{ife} is denoted by $F_G$, $F_L$ and $F_I$ respectively.

We initialized the weights of the \glspl{fe} with the pretrained model found in PyTorch framework \cite{paszke2019pytorch} that is initially trained for classification problem with the ImageNet dataset \cite{deng2009imagenet}. For \gls{ife}, we initialize the weights of all Informed Convolution layers with their corresponding `normal' convolution layers found in the pretrained model. The reason is that a `normal' convolution layer is a special case of the Informed Convolution layer and it can be a good initialization for the weights of \gls{ife}. However, we found that the difference in performance is insignificant between with and without the pretrained initialization of the \gls{ife}.

\subsection{Object Descriptor}\label{subsec:supp_descriptor}

\begin{table*}\centering\caption{Structure of Object Descriptor ($\mathcal{D}$)}\label{tab:descriptor_structure}
\resizebox{\textwidth}{!}{
\begin{tabular}{l|cccccc}\Xhline{4\arrayrulewidth}
Layer Name  & $\mathcal{D}$.C.1  & $\mathcal{D}$.C.2 & $\mathcal{D}$.M.1 & $\mathcal{D}$.M.2 & $\mathcal{D}$.P.1 & $\mathcal{D}$.P.2 \\\hline
Input & $\hat{\mathbf{c}}$ & $\mathcal{D}$.C.1 &  $M$ & $\mathcal{D}$.M.1 & $\mathcal{D}$.M.2 $\bigotimes$ $\mathcal{D}$.C.2 & $\mathcal{D}$.P.1 \\
Structure & $\begin{bmatrix}\text{Linear}(2260, 512)\\\text{relu}\end{bmatrix}$ & $\begin{bmatrix}\text{Linear}(512, 32)\\\text{softmax}\end{bmatrix}$ & $\begin{bmatrix}3\times3~\text{conv}(1, 16)/2\\\text{batch norm}\\\text{relu}\end{bmatrix}$ &  $\begin{bmatrix}3\times3~\text{conv}(16, 32)/2 \\\text{batch norm}\\\text{relu}\end{bmatrix}$ &  $\begin{bmatrix}3\times3~\text{conv}(32, 64)\\\text{batch norm}\\\text{relu}\\\text{upsample}(2)\end{bmatrix}$ &  $\begin{bmatrix}3\times3~\text{conv}(64, 128)\\\text{batch norm}\\\text{sigmoid}\\\text{upsample}(2)\end{bmatrix}$ \\
Output & $512$ & $32$ & $16\times112\times112$ & $32\times56\times56$ & $64\times112\times112$ & $128\times224\times224$\\
& & & & & & $D$ \\\hline
\end{tabular}%
}
\end{table*}

The Object Descriptor, \Cref{tab:descriptor_structure}, is depicted in \Cref{fig:object_descriptor}. It has two input branches for the category embedding and the binary mask of the object. The category embedding branch $\mathcal{D}.C$ consists of two fully connected layers while the binary mask branch $\mathcal{D}.M$ consists of two 2D-Convolution layers. The output of $\mathcal{D}.C$ is broadcasted and multiplied with $\mathcal{D}.M$ as in \Cref{eq:obj_desc_intermediate}.

We have investigated the usage of the cropped image $I_C$ in the generation of the description $D$. However, we noticed that it actually increases the complexity while not increasing the performance. Since all important features are extracted from $I \& I_C$ by our strong \glspl{fe}. The object descriptor needs only to `learn' where to give attention. This can mainly be obtained by information about 1) where the object is located in the input image $I_C$ which is provided through the binary mask $M$, and 2) what category this object belongs to which is provided by the self-learned category embedding $\hat{\mathbf{c}}$.

\subsection{Gating Mechanism}\label{subsec:supp_gate}

\begin{table*}\centering\caption{Structure of Gates ($\mathcal{G}$)}\label{tab:gate_structure}
\resizebox{\linewidth}{!}{
\begin{tabular}{l|ccc}\Xhline{4\arrayrulewidth}
Layer Name  & $\mathcal{G}$.1  & $\mathcal{G}$.2 & $\mathcal{G}$.3 \\\hline
Input & $D$ & $\mathcal{G}$.1 & $\mathcal{G}$.2 \\
Structure & $\begin{bmatrix}3\times3~\text{conv}(128, 64)/2\\\text{batch norm}\\\text{relu}\end{bmatrix}$ &  $\begin{bmatrix}3\times3~\text{conv}(64, 32)/2 \\\text{batch norm}\\\text{relu}\end{bmatrix}$ & $\begin{bmatrix}3\times3~\text{conv}(32, 3)/2 \\\text{batch norm}\\\text{sigmoid}\end{bmatrix}$  \\
Output & $64\times112\times112$ & $32\times56\times56$ & $3\times28\times28$ \\
& & & $A_G, A_L$ or $A_I$ \\\hline
\end{tabular}%
}
\end{table*}

We have three gates $\mathcal{G}_G, \mathcal{G}_L \& \mathcal{G}_I$ for the three different extracted features $F_G, F_L \& F_I$ and their corresponding outputs are $A_G, A_L \& A_I$, respectively. \Cref{tab:gate_structure} shows the architecture of each gate. The input is $D$ (the description of the object), which is the output of $\mathcal{D}$ (the Object Descriptor). We use a final Sigmoid activation function to ensure that the output range of the learned attention is between $0$ and $1$. This guarantees the numerical stability of the network, as later the produced values $A_G, A_L \& A_I$ are multiplied with the features $F_G, F_L \& F_I$. Without bounding the range of the learned attention maps, the values may explode.

\subsection{Interpreter}\label{subsec:supp_interpreter}

\begin{table*}\centering\caption{Structure of Interpreter ($\mathcal{I}$)}\label{tab:interpreter_structure}
\resizebox{\textwidth}{!}{
\begin{tabular}{l|cccc}\Xhline{4\arrayrulewidth}
Layer Name  & $\mathcal{I}$.E.1  & $\mathcal{I}$.E.2 & $\mathcal{I}.\text{H}_i$.1 & $\mathcal{I}.\text{H}_i$.2 \\\hline
Input & $f_T$ & $\mathcal{I}$.E.1 & $\mathcal{I}$.E.2 & $\mathcal{I}.\text{H}_i$.1 \\
Structure & $\begin{bmatrix}\text{Linear}(768, 512)\\\text{relu}\end{bmatrix}$ & $\begin{bmatrix}\text{Linear}(512, 256)\\\text{softmax}\end{bmatrix}$ & $\begin{bmatrix}\text{Linear}(256, 128)\\\text{relu}\end{bmatrix}$ & $\begin{bmatrix}\text{Linear}(128, 620)\\\end{bmatrix}$ \\
Output & $512$ & $256$ & $128$ & $620$ \\
 & & & & $\hat{\mathbf{a}}$ \\\hline
\end{tabular}%
}
\end{table*}

The interpreter, \Cref{tab:interpreter_structure}, consists of two stages as shown in \Cref{fig:interpreter}. The first part $\mathcal{I}.E$ reduces the length of the feature vector to $256$. Then for each category, we have its own $\mathcal{I}.\text{H}_i, \forall i\in \{1,2,\cdots c\}$, where $c$ is the number of categories. The output length in the case of \gls{vaw} is $620$ as the set of attributes is fixed over all categories. However, the output length in the case of \gls{car} varies depending on the set of possible attributes of each category. For example, the output length for a Pedestrian object would be $38$. While the output length for a Mid-to-Large Vehicle is $41$. Please refer to \Cref{sec:exp_res,sec:supp_disc_data} for more details and discussion about the datasets.

\begin{table*}\centering\caption{Structure of Category Estimator ($\mathcal{C}$) and local and intrinsic attributes estimators (LAE \& IAE)}\label{tab:ce_ae_structure}
\resizebox{\textwidth}{!}{
\begin{tabular}{l|c||c||c}\Xhline{4\arrayrulewidth}
Layer Name  & $\mathcal{C}$.1  & LAE.1 & IAE.1 \\\hline
Input & $F_L$ & $F_L$ & $F_I$ \\
Structure & $\begin{bmatrix}\text{Linear}(768, 2260)\end{bmatrix}$ & $\begin{bmatrix}\text{Linear}(768, 620)\end{bmatrix}$ & $\begin{bmatrix}\text{Linear}(768, 620)\end{bmatrix}$ \\
Output & $2260$ & $620$ & $620$ \\
 & $\hat{\mathbf{c}}$ & $\hat{\mathbf{a}}_l$ & $\hat{\mathbf{a}}_i$ \\\hline
\end{tabular}%
}
\end{table*}

\subsection{Category Estimator}\label{subsec:supp_category}

The first column of \Cref{tab:ce_ae_structure} shows the structure of the category estimator, which is a single fully connected layer with an output length equal to the number of categories. Ideally, the Category Estimator should produce a one-hot encoding representation of the category of the object. However, practically it produces a Probability Mass Function (PMF) of the object's category. If the PMF has two or more peaks, then that is primarily due to the visual similarity between their corresponding categories. In other words, the generated embedding captures visual similarities between different categories based on the shape of the object of interest. As we have argued in \Cref{tab:category_ablation}, this is better than a fixed pretrained word embedding such as GloVe \cite{pennington2014glove} in Pham \etal \cite{pham2021learning}.

\subsection{Multi-Object Detection Head}\label{subsec:supp_modh}

\begin{table*}\centering\caption{Structure of Multi-Object Detection Head (MODH)}\label{tab:modh_structure}
\resizebox{\textwidth}{!}{
\begin{tabular}{l|ccc}\Xhline{4\arrayrulewidth}
Layer Name  & MODH.1  & MODH.2 & MODH.3 \\\hline
Input & $F_G$ & MODH.1 &  MODH.2 \\
Structure & $\begin{bmatrix}3\times3~\text{conv}(768, 512)\\\text{batch norm}\\\text{relu}\end{bmatrix}$ & $\begin{bmatrix}3\times3~\text{conv}(512, 256)\\\text{batch norm}\\\text{relu}\end{bmatrix}$ & $\begin{bmatrix}3\times3~\text{conv}(256, 2265)\\\text{batch norm}\\\gamma\end{bmatrix}$\\
Output & $512\times28\times28$ & $256\times28\times28$ & $2265\times28\times28$ \\ 
 & & & $\hat{O}_G$ \\\hline
\end{tabular}%
}
\end{table*}

The multi-object-detection head (MODH), \Cref{tab:modh_structure}, detects different objects in the whole given image. The output should not change by using different objects in the same image as the input of \gls{gfe} is the whole input image $I$. The input to the MODH is the upsampled and concatenated features ($F_G$) \Cref{eq:fg}. The output has $2265$ channels in case of using \gls{vaw} dataset and $17$ in case of using \gls{car}, since the number of categories is different in each dataset. The final output passes by a custom activation function $\gamma$ that splits the channels of the input features into sets and passes each set to a different activation function depending on the purpose of this set. $\gamma$ is defined as follows. For the first channel that represents the confidence, the activation is a Sigmoid as well as for the bounding box center coordinates. On the other hand, the activation is an exponential for the dimensions of the bounding box (to ensure non-negativity) while it is a softmax for the multi-category channels (to ensure PMF axioms -- non-negativity and summation to one).

\subsection{Mask Estimator}\label{subsec:supp_mask}

\begin{table*}\centering\caption{Structure of Mask Estimator ($\mathcal{M}$)}\label{tab:mask_estimator_structure}
\resizebox{\textwidth}{!}{
\begin{tabular}{l|cccc}\Xhline{4\arrayrulewidth}
Layer Name  & $\mathcal{M}$.1  & $\mathcal{M}$.2 & $\mathcal{M}$.3 & $\mathcal{M}$.4 \\\hline
Input & $F_L$ & $\mathcal{M}$.1 &  $\mathcal{M}$.2 & $\mathcal{M}$.3 \\
Structure & $\begin{bmatrix}3\times3~\text{conv}(768, 256)\\\text{batch norm}\\\text{relu}\\\text{upsample}(2)\end{bmatrix}$ & $\begin{bmatrix}3\times3~\text{conv}(256, 128)\\\text{batch norm}\\\text{relu}\\\text{upsample}(2)\end{bmatrix}$ & $\begin{bmatrix}3\times3~\text{conv}(128, 64)\\\text{batch norm}\\\text{relu}\\\text{upsample}(2)\end{bmatrix}$ &  $\begin{bmatrix}3\times3~\text{conv}(64, 1)\\\text{batch norm}\\\text{sigmoid}\end{bmatrix}$\\
Output & $256\times56\times56$ & $128\times112\times112$ & $64\times224\times224$ & $1\times224\times224$ \\
& & & & $\hat{M}$\\\hline
\end{tabular}%
}
\end{table*}

The Mask Estimator ($\mathcal{M}$) is a temporary decoder that takes the output features of \gls{lfe} ($F_L$) \Cref{eq:fl}. Its structure is depicted in \Cref{tab:mask_estimator_structure}. The output is a single channel representing the estimation of the binary mask. The output is restricted to be between $0$ and $1$ through a final Sigmoid layer -- one indicates the pixel belongs to the object while zero indicates that this pixel does not belong to the object of interest.

\subsection{Attribute Predictors}\label{subsec:supp_attributes_1}

The second and third column of \Cref{tab:ce_ae_structure} shows the structure of the local and intrinsic temporary decoders respectively. They both have the same structure, however, the inputs and outputs of each are distinct. In the case of LAE, the inputs and outputs are $F_L$ \Cref{eq:fl} and $\mathbf{\hat{a_l}}$ respectively. While for IAE, the inputs and outputs are $F_I$ \Cref{eq:fi} and $\mathbf{\hat{a_i}}$ respectively.

\section{Discussion of Datasets}\label{sec:supp_disc_data}

In \Cref{sec:exp_res}, we have trained \gls{glidenet} using two new and challenging datasets \gls{vaw} \cite{pham2021learning} and \gls{car} \cite{metwaly2022car}. 
The following are a few of the reasons why these two datasets were chosen in particular: 
\begin{enumerate}
    \item the number and diversity of categories in both datasets are high enough to evaluate the validity and effectiveness of the proposed multi-category architecture.
    
    \item \gls{vaw} has already been used in attributes prediction as in \cite{pham2021learning}. 
    
    \item \gls{car} has a complex taxonomy with different sets of attributes depending on different categories. Furthermore, unlike \gls{vaw}, the attributes in \gls{car} are not binary or ternary. This makes \gls{car} more challenging as well as adds flavor to the comparison of \gls{glidenet} with other methods.
\end{enumerate}

\begin{figure}
    \centering
    \includegraphics[height=0.95\textheight]{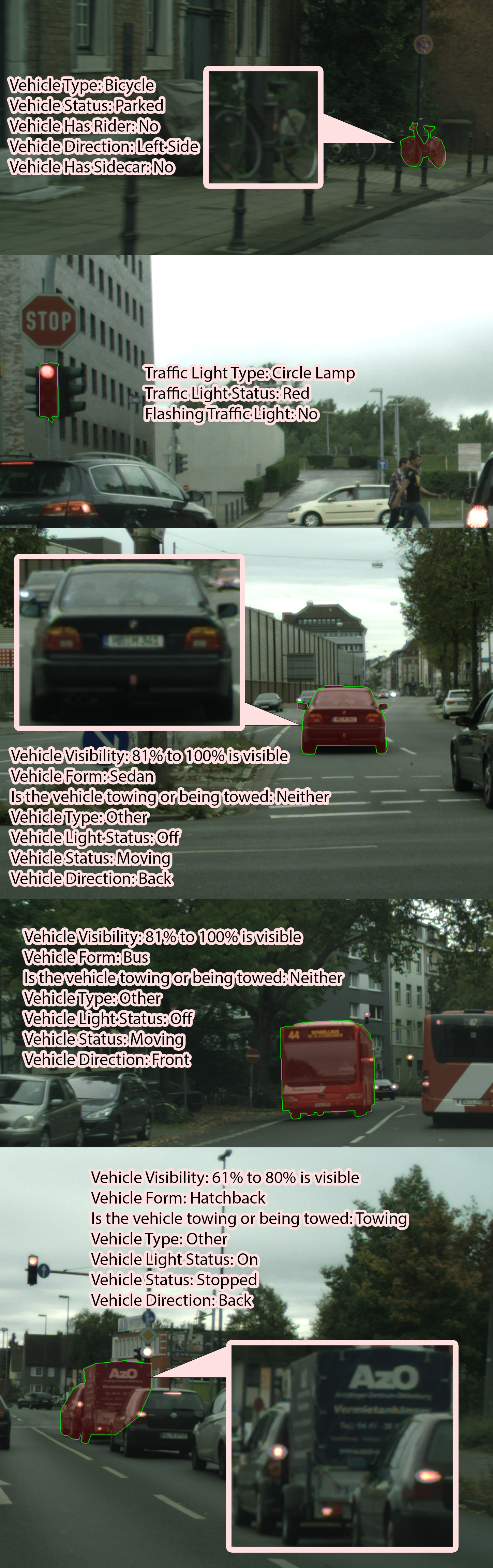}
    \caption{Examples of the \gls{car} Dataset. Figure from the original paper \cite{metwaly2022car}.}
    \label{fig:car_example}
\end{figure}
\begin{figure}
    \centering
    \includegraphics[width=\linewidth]{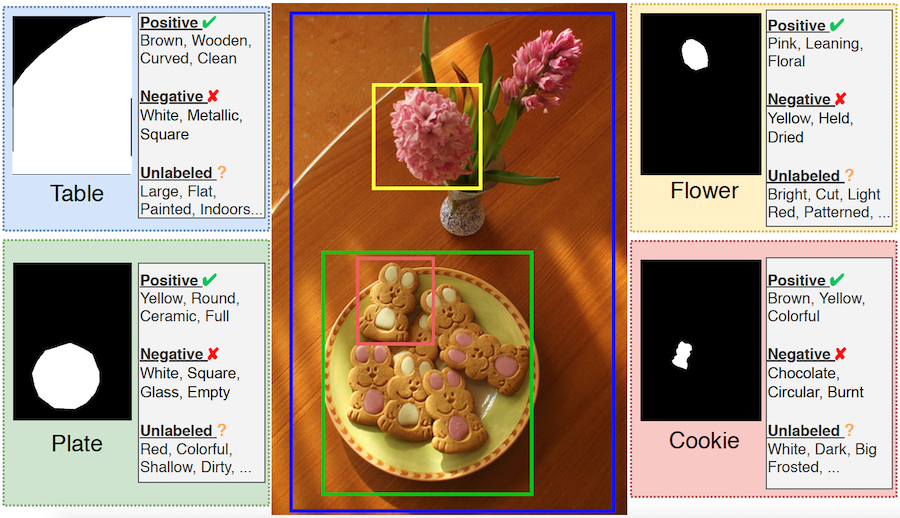}
    \caption{Examples of the \gls{vaw} Dataset. Figure from the original paper \cite{pham2021learning}}
    \label{fig:vaw_example}
\end{figure}

\Cref{fig:car_example,fig:vaw_example} show examples from the original papers of both datasets. \gls{car}\footnote{An API is provided at \url{https://github.com/kareem-metwaly/CAR-API}} focuses on the application of attributes prediction for self-driving vehicles. Therefore, \gls{car} focus on attributes such as the activity of a pedestrian, visibility of a vehicle, the color of a traffic light, the speed limit of a traffic sign. On the other hand, \gls{vaw}\footnote{The authors provide the dataset through their website \url{http://vawdataset.com/}} is pretty generic and has a wider variety of categories but it has the same set of attributes for all. Most of those attributes are unlabeled; since the majority of them are not meaningful to a certain category. All attributes in \gls{vaw} can take one of three different labels; positive, negative, and unlabeled.

\subsection{Objects with low pixel-count:}  
\begin{figure}
    \centering
    \begin{subfigure}{0.24\linewidth}
    \includegraphics[width=\textwidth]{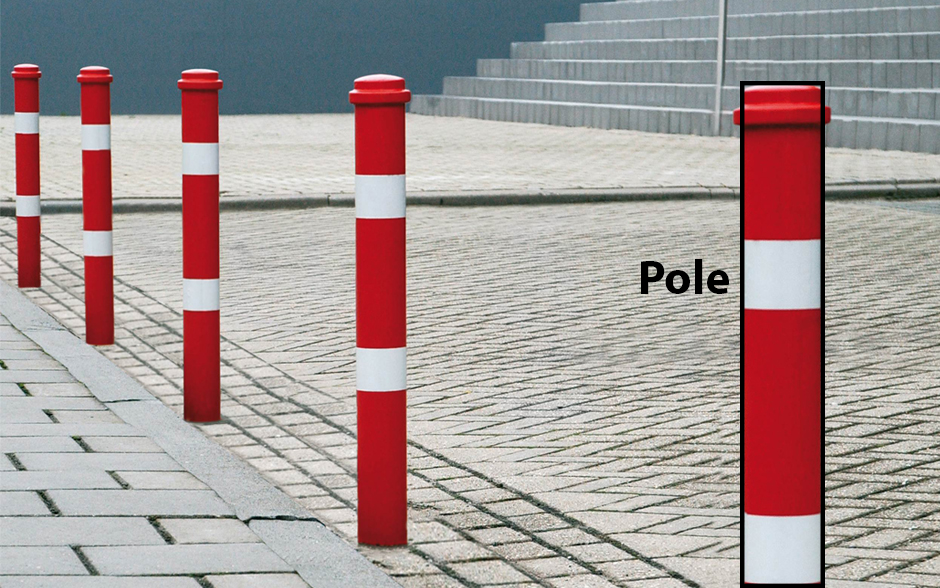}
    \caption{The pole is the object of interest}
    \end{subfigure}\hfill
    \begin{subfigure}{0.2\linewidth}
    \includegraphics[width=\textwidth]{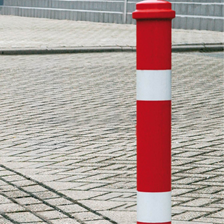}
    \caption{Cropped}
    \end{subfigure}\hfill
    \begin{subfigure}{0.2\linewidth}
    \includegraphics[width=\textwidth]{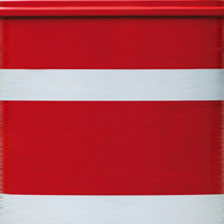}
    \caption{Stretched}
    \end{subfigure}\hfill
    \begin{subfigure}{0.3\linewidth}
    \includegraphics[width=\textwidth]{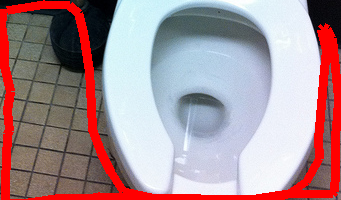}
    \caption{The floor is the object of interest}
    \label{fig:floor_example}
    \end{subfigure}
    \caption{Objects that have benefited from IFE}
    \label{fig:ife_importance}
\end{figure}

\Cref{fig:ife_importance} depicts two examples that demonstrate the importance of IFE. Let's look at the first example (a), (b) \& (c), a narrow vertical pole. To predict its attributes without using IFE, we can either crop while keeping the aspect ratio (no distortion) or stretch it (distorting the image). And both techniques influence the prediction of other attributes. For example, image (c) no longer looks like a pole. Additionally, (d) displays another example (from the VAW dataset) where cropping and stretching are difficult due to the toilet being surrounded by the floor. IFE would easily help these cases, allowing information to flow only from pixels of interest.

\section{Experimental Setup}\label{sec:supp_exp_setup}

In our Experiments and Results \Cref{sec:exp_res}, we have compared \gls{glidenet} with four different state-of-the-art methods \cite{durand2019learning,jiang2020defense,sarafianos2018deep,pham2021learning}. We also performed ablation study to prove the importance and effectiveness of the different \glspl{fe} as well as the novel convolution layer -- Informed Convolution.

In our training, we have trained the network at Stage I for 15 epochs, then we switched to Stage II for 10 epochs. We have noticed that changing the number of epochs slightly for each stage did not have a noticeable change in performance. The temporary decoders are removed during Stage II and inference stage; they are only used in Stage I to guide the \glspl{fe} for their supposed objectives. As mentioned in \Cref{sec:exp_res}, we set the values of the hyperparameters of the training loss function to be as follow: $\lambda_{gp0} = 1$, $\lambda_{gp} = 0.01$, $\lambda_{gd} = 0.5$, $\lambda_{gc} = 0.5$, $\lambda_{lm} = 0.1$, $\lambda_{lc} = 0.01$, $\lambda_{la} = 1$, $\lambda_{ia} = 1$ and $\lambda_{lc2} = 0.01$. We have attempted training with different values. What we noticed the most is that it is important to keep the values of $\lambda_{gp}, \lambda_{lc} \& \lambda_{lc2}$ lower than other values significantly. Otherwise the training diverges and the training focuses more on estimating the correct category than actually ensuring decent performance for other decoders.

We have developed our code based on the PyTorch framework \cite{paszke2019pytorch}. We used GPUs to speed up the training, specifically, we used NVIDIA Tesla V100 \textregistered  GPUs. We distributed the code over 4 GPUs to speed up the training. The training time was less than one day and the average inference time per an entire image was $\sim 0.05$ second. This is a very reasonable time for a real-time application. Typically, it is not required to predict attributes with a frequency greater than $20$ Hz in most applications such as autonomous vehicles. For an average speed of $60$ MPH ($88$ feet/second), a vehicle will on average predict attributes of the scene every 4 to 5 feet ($88/20$). This is a small traveling distance for the scene to change significantly. In other words, $20$ Hz is sufficient to estimate attributes of all objects in a scene and make fast real-time decisions based on the predicted attributes.
It is worth noting that this time can significantly be reduced per scene if we kept the produced General Features $F_G$ from one instance to another in the same scene (image); as they all share the same scene. In addition, the time can be reduced by tracking objects with predicted attributes. This will help in decreasing the number of objects that require attributes prediction in each scene, which in turn decreases the computation time per scene.

{\small
\bibliographystyle{./ieee_fullname}
\bibliography{./main}
}

\end{document}